\documentclass[runningheads]{llncs}
\usepackage{graphicx}
\usepackage{amsfonts}		% Additional math fonts
\usepackage{amsmath}		% Math symbols
\usepackage{amssymb}
\usepackage{latexsym}
\DeclareMathOperator*{\argmin}{argmin}
\DeclareMathOperator*{\argmax}{argmax}

\begin{document}
\title{Learning to Segment Medical Images with Scribble-Supervision Alone}
\titlerunning{Learning to Segment Medical Images with Scribble-Supervision Alone}

\author{Yigit B. Can\inst{1} \and 
Krishna Chaitanya\inst{1} \and 
Basil Mustafa\inst{1,3}\thanks{Basil Mustafa contributed to this work during a research internship at ETH.} \and
Lisa M. Koch\inst{2} \and
Ender Konukoglu\inst{1} \and
Christian F. Baumgartner\inst{1}}

\authorrunning{Y.B. Can et al.}

\institute{Computer Vision Lab, ETH Z\"urich, Switzerland \and
Computer Vision and Geometry Group, ETH Z\"urich, Switzerland \and 
University of Cambridge, UK}
\maketitle          
\begin{abstract}

Semantic segmentation of medical images is a crucial step for the quantification of healthy anatomy and diseases alike. The majority of the current state-of-the-art segmentation algorithms are based on deep neural networks and rely on large datasets with full pixel-wise annotations. Producing such annotations can often only be done by medical professionals and requires large amounts of valuable time. Training a medical image segmentation network with weak annotations remains a relatively unexplored topic. In this work we investigate training strategies to learn the parameters of a pixel-wise segmentation network from scribble annotations alone. We evaluate the techniques on public cardiac (ACDC) and prostate (NCI-ISBI) segmentation datasets. We find that the networks trained on scribbles suffer from a remarkably small degradation in Dice of only 2.9\% (cardiac) and 4.5\% (prostate) with respect to a network trained on full annotations. 

\end{abstract}

\section{Introduction}
\label{sec:introduction}

Convolutional neural networks (CNN) have been used for semantic segmentation on medical images with great success \cite{bernard2018deep}. For the most part, these methods rely on fully annotated images to train the network. Although CNN-based segmentation algorithms keep evolving and improving, the amount of available training data still has a substantial effect on the performance \cite{DBLP:conf/cvpr/LinDJHS16}. However, it is difficult to obtain large scale fully annotated data for medical images since it requires an expert to spend considerable time and effort. 

To address this limitation, a number of works have proposed interactive image segmentation methods relying on weak annotations such as bounding boxes \cite{DBLP:journals/tog/RotherKB04}, or scribbles \cite{criminisi2008geos,grady2006random}. However, in these works, the annotations need to be provided for each new test image. Recently, a number of works have demonstrated that it is feasible to train fully-automatic, learning-based algorithms using exclusively weak labels \cite{DBLP:conf/cvpr/LinDJHS16,DBLP:conf/iccv/DaiHS15,DBLP:journals/corr/PapandreouCMY15,rajchl2017deepcut}. Despite being trained on weak labels, these methods can produce full segmentation masks on test images. Of the above works only \cite{rajchl2017deepcut} was demonstrated on medical images. The authors proposed to train a segmentation network for fetal structures from bounding box annotations only. 

\begin{figure}[t]

\centering
\includegraphics[width=1\textwidth ]{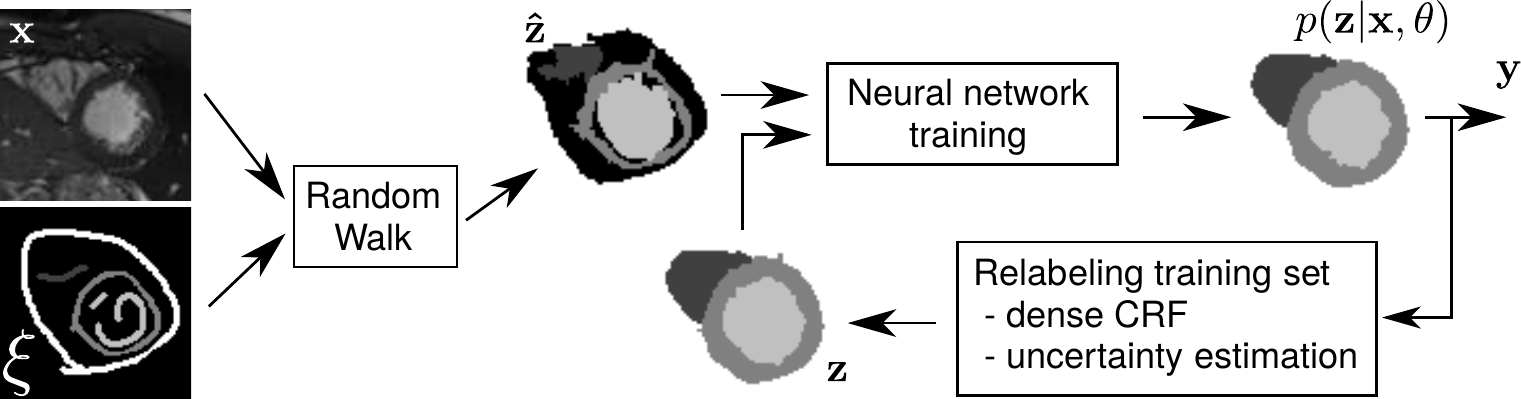}
\caption{Overview of the proposed training framework.}
\label{fig:process}

\end{figure}

In this paper we present a scribble-based weakly-supervised learning framework for medical images. Scribbles have been recognized as particularly user-friendly form of supervision \cite{DBLP:conf/cvpr/LinDJHS16} and may be better suited for nested structures, when compared to bounding boxes. Furthermore, they require only a fraction of the annotation time compared to full pixel-wise annotations. Following previous works, the proposed framework is an iterative two-step procedure in which a segmentation network is trained on the scribble annotations, then this network is used in conjunction with a conditional random field (CRF) to relabel the training set. This in turn is used for an additional training \emph{recursion}\footnote{We refer to this as recursion rather than iteration to avoid confusion with single mini-batch gradient descent steps, which are also often referred to as iteration.}. We show that this procedure, under some assumptions, can be interpreted as expectation maximization (EM). We investigate multiple strategies for relabeling the training dataset, estimating the CRF parameters, and quantifying uncertainty in the relabeling step. An overview of the method is shown in Fig. \ref{fig:process}. %, which occur frequently in medical images.

We evaluate the framework and its individual components on the public cardiac ACDC dataset \cite{bernard2018deep} and the NCI-ISBI 2013 prostate segmentation challenge data \cite{bloch2015nci}. We show that despite the inherently very sparse nature of the annotations the proposed methods achieve a segmentation accuracy within 95\% of a baseline network trained with full supervision. To our knowledge, this is the first demonstration of training a pixel-wise segmentation network with scribble supervision on medical image data. 

\section{Methods}

\begin{figure}[t]

\centering
\includegraphics[width=0.90\textwidth ]{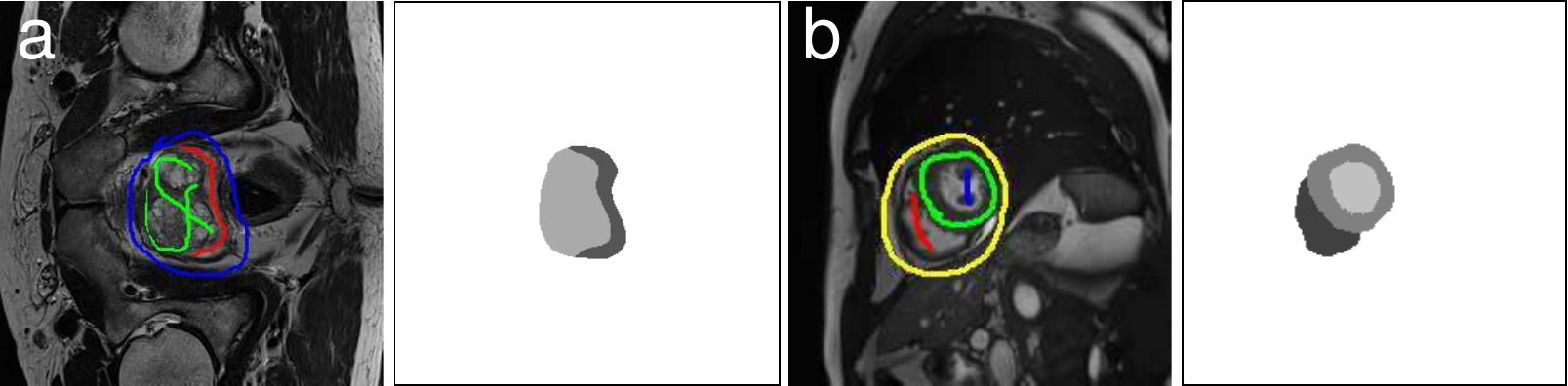}
\caption{Example images and scribbles on the left and ground truth segmentations on the right for the a) prostate and b) cardiac datasets, respectively.}
\label{fig:scribbles}
\end{figure}

The aim of our proposed method is to learn the parameters $\theta$ of a CNN-based segmentation network $\mathbf{y} = f(\mathbf{x}; \theta)$ such that it predicts a generally unknown segmentation mask $\mathbf{y} \in \{0,\ldots,L\}^N$ for an input image $\mathbf{x}\in\mathbb{R}^N$, where $N$ is the number of pixels. During training, rather than full pixel-wise annotations, we are only provided with a ground truth annotation $\mathbf{\xi}$ for a small number of pixels (i.e. the scribbles). Note that this also includes a background scribble (see examples in Fig. \ref{fig:scribbles}). The proposed framework consists of a repeated estimation of the network parameters and subsequent relabeling of the training dataset by combining the network prediction with a CRF. We investigate two different CRF inference strategies: the dense CRF approach proposed in \cite{krahenbuhl2011efficient}, and a recent extension thereof in which the CRF is formulated as a recurrent neural network (RNN) and the CRF parameters can be learned end-to-end \cite{DBLP:conf/iccv/0001JRVSDHT15}. Moreover, we investigate a novel strategy for incorporating prediction uncertainty in the relabeling step based on \cite{DBLP:journals/corr/KendallBC15}. For all investigated strategies we perform an initial region growing step described in the following.

\subsection{Generation of Seed Areas by Region Growing}

For this step we use the random walk-based segmentation method proposed by \cite{grady2006random}, which (similar to neural networks) produces a pixel-wise probability map for each label. We assign each pixel its predicted value only if the probability exceeds a threshold $\tau$. Otherwise the pixel-label is treated as \emph{unknown}. An example of this step can be seen in Fig. \ref{fig:process}. Note that the threshold is intentionally chosen very high such as to underestimate the true extent of the structures and only include pixels which have a very high probability of being correctly estimated. Those assignments will serve as new ``ground truth'' labels $\mathbf{\hat{z}}$ for the remainder of the steps and will be referred to as seed areas. The uncertain pixels $\mathbf{z}$ are treated as unlabeled, i.e. they are the latent variables of our model. %Note that this region growing step is not essential, however, in preliminary experiments we found that it leads to faster convergence and superior results. of \textcolor{red}{0.99 for cardiac and 0.90 for prostate}

\subsection{Separate CRF and Network Training}

We propose a hard expectation maximization (EM) approximation to learn the network parameters $\theta$ in an iterative fashion. The algorithm consists of alternatingly estimating the best parameters of the neural network given a labeling obtained using the current parameters $\theta^{old}$ (M step), and estimating the optimal labeling of the latent variables given an updated $\theta$ (E step). We assume the following graphical model 
\begin{equation}\label{eq:graphical_model}
p(\mathbf{z}, \mathbf{\hat{z}} | \mathbf{x}, \theta) = p(\mathbf{z} | \mathbf{x}, \theta)p(\mathbf{\hat{z}}|\mathbf{z}, \mathbf{x}),
\end{equation}
where $p(\mathbf{z | \mathbf{x}, \theta})$ is modeled using a neural network $f(\mathbf{x};\theta)$. Following the standard EM approach, we write the expectation of the complete-data log likelihood as
\begin{equation}\label{eq:Q}
Q(\theta, \theta^{old}) = \sum_\mathbf{z} p(\mathbf{z} | \mathbf{\hat{z}}, \mathbf{x}, \theta^{old}) \ln p(\mathbf{z}, \mathbf{\hat{z}} | \mathbf{x}, \theta).
\end{equation}
In the E step of the algorithm we estimate the mode of $ p(\mathbf{z} | \mathbf{\hat{z}}, \mathbf{x}, \theta^{old}) $ as
% \begin{equation}\label{eq:opt_z}
% \begin{split}
% \mathbf{z^* } & = \argmax_\mathbf{z} p(\mathbf{z} | \mathbf{\hat{z}}, \mathbf{x}, \theta^{old}) = \argmax_\mathbf{z} \frac{p(\mathbf{z}, \mathbf{\hat{z}} | \mathbf{x}, \theta^{old})}{p(\mathbf{\hat{z}|\mathbf{x}})} \\
% 			  & = \argmax_\mathbf{z} p(\mathbf{z} , \mathbf{\hat{z}} | \mathbf{x}, \theta^{old}),
% \end{split}
% \end{equation}
\begin{equation}\label{eq:opt_z}
\mathbf{z^* } = \argmax_\mathbf{z} p(\mathbf{z} | \mathbf{\hat{z}}, \mathbf{x}, \theta^{old}) = \argmax_\mathbf{z} \frac{p(\mathbf{z}, \mathbf{\hat{z}} | \mathbf{x}, \theta^{old})}{p(\mathbf{\hat{z}|\mathbf{x}})}
			  = \argmax_\mathbf{z} p(\mathbf{z} , \mathbf{\hat{z}} | \mathbf{x}, \theta^{old}),
\end{equation}
using the fact that $p(\mathbf{\hat{z}|\mathbf{x}})$ does not depend on $\mathbf{z}$. 

By assuming a complete dependency graph between all $\mathbf{z},\mathbf{\hat{z}}$, the conditional joint distribution can be factorized and the E step can be written as the following CRF optimization problem:
\begin{equation}\label{eq:CRF}
\begin{split}
\mathbf{z^* } & = \argmin_\mathbf{z} \sum_{i \in \mathcal{C}_u(\mathbf{z})} \psi_u(z_i | \mathbf{x}, \theta^{old}) +
\sum_{i \in \mathcal{C}_u(\mathbf{\hat{z}})} \hat{\psi}_{u}(\hat{z}_i) + \\
& \sum_{i,j \in \mathcal{C}_p(\mathbf{z})} \psi_p(z_i, z_j | x_i, x_j) +
\sum_{i,j \in \mathcal{C}_p(\mathbf{\hat{z}})} \psi_p(\hat{z}_i, \hat{z}_j | x_i, x_j) +
\sum_{i,j \in \mathcal{C}_p(\mathbf{z}, \mathbf{\hat{z}})} \psi_p(z_i, \hat{z}_j | x_i, x_j), 
\end{split}
\end{equation} 
where $\mathcal{C}_u(\cdot)$ denotes the set of all unary cliques of a set of variables and $\mathcal{C}_p(\cdot)$ denotes the set of all pairwise cliques. The unary potential function $\psi_u$ acting on the latent variables is defined using the current network output as
\begin{equation}\label{eq:unary}
\psi_u(z|\mathbf{x}, \theta^{old}) = -\ln p(z_i | \mathbf{x}, \theta^{old}) = -\ln f(x;\theta^{old}).
\end{equation}
The unary potential function $\hat{\psi}_u$ acting on the seed regions $\mathbf{\hat{z}}$ is defined as 0 for labellings matching the ground truth and infinity otherwise, effectively preventing the initially grown regions from changing. Furthermore, we use the pairwise potential function $\psi_p$ proposed in \cite{krahenbuhl2011efficient}:
\begin{equation}\label{eq:pairwise}
\begin{split}
\psi_p(z_i, z_j |  x_i, x_j) = \mu(z_i, z_j) & \left( w_1\exp\left(-\frac{dist(x_i, x_j)^2}{2\sigma_\alpha^2} -\frac{|x_i - x_j|^2}{2\sigma_\beta^2} \right) + \right. \\ 
							 					    & \left. w_2\exp\left(-\frac{dist(x_i, x_j)^2}{2\sigma_\gamma^2} \right) \right),
\end{split}
\end{equation}
where the label compatibility function is given by the Potts model $\mu(z_i, z_j) = [z_i \neq z_j]$, and $dist(\cdot, \cdot)$ denotes the Euclidean distance between the pixel locations. We estimate the hyperparameters $w_1, w_2, \sigma_\alpha, \sigma_\beta, \sigma_\gamma$ in a grid search on the validation set. In order to optimize Eq. \ref{eq:CRF} we use the approach in \cite{krahenbuhl2011efficient}. We also consider a simple modification of this procedure as a baseline in which we set the pairwise terms to zero and only use the unary terms $\psi_u, \hat{\psi}_u$. 

In the M step, after we have found the optimal labeling of the latent variables $\mathbf{z^*}$ using the network parameters $\theta^{old}$ we can rewrite Eq. \ref{eq:Q} as 
\begin{equation}
\begin{split}
Q(\theta, \theta^{old}) & \approx \sum_{\mathbf{z}}\delta(\mathbf{z}=\,\mathbf{z^*}|\mathbf{\hat{z}},\mathbf{x},\theta^{old}) \ln p(\mathbf{z},\mathbf{\hat{z}}| \mathbf{x}, \theta) \\
                        & = \ln p(\mathbf{z^*} | \mathbf{x}, \theta) + \ln p(\mathbf{\hat{z}}|\mathbf{z^*}, \mathbf{x}),
\end{split}
\end{equation}
where $\delta$ is the Dirac delta function, the approximate equality is due to the hard EM approximation and we substituted Eq. \ref{eq:graphical_model} to obtain the equality. 
Since $\ln p(\mathbf{\hat{z}}|\mathbf{z}, \mathbf{x})$ does not depend on $\theta$ the optimization can be written as 
\begin{equation}
\theta^* = \argmax_\theta \ln p(\mathbf{z^*} | \mathbf{x}, \theta).
\end{equation}
We find the parameters $\theta$ that maximize the likelihood of predicting the labels $\mathbf{z^*}$ by minimizing the pixel-wise cross entropy function between the labels $\mathbf{z^*}$ and the network output using the ADAM optimizer with an initial learning rate of 0.001 which is multiplied by 0.9 every 3000 iterations. We use the modified U-Net segmentation network used in \cite{DBLP:conf/miccai/BaumgartnerKPK17} in all experiments. The network parameters $\theta$ for each recursion are initialized with $\theta^{old}$. The E and the M steps get repeated until convergence, which typically occurs within 3 recursions or less. 

In the first recursion, we set the cross-entropy loss to zero in all locations where the random walk is ``uncertain'' (probabilities below $\tau$), allowing the network to predict any label in those regions. We also explore a strategy to identify uncertain regions in subsequent iterations, which will be discussed in Sec.\,\ref{sec:uncertainty}

\subsection{Integrated Network Training and  (CRF-RNN)}

Here, we investigate estimation of the CRF parameters as part of the network training. To that end we use the CRF-RNN layer proposed in \cite{DBLP:conf/iccv/0001JRVSDHT15} which learns individual kernel weights for each class and a more flexible compatibility matrix. %This gives the CRF more flexibility and makes it less sensitive to the choice of the hyperparameters. %The standard deviations $\sigma_\alpha, \sigma_\beta, \sigma_\gamma$ in Eq. \ref{eq:pairwise}, are non-learnable parameters, which we set to the same values as in \cite{DBLP:conf/iccv/0001JRVSDHT15}. 

To obtain a new labeling $\mathbf{z^*}$ we simply run a forward pass through the network. Next, in order to prevent the original seed regions $\mathbf{\hat{z}}$ from changing, we simply reset those values to their original label. In future work, we aim to include this constraint directly into the CRF-RNN formulation. %In the first step we evaluate the new labeling $\mathbf{z^*}$ by simply running a forward pass through the network including the CRF-RNN layer.

In the subsequent network optimization step, we directly learn to predict those $\mathbf{z^*}$. Here we use the following training scheme: the network parameters are trained as above for 10 mini-batch iterations while keeping the RNN parameters constant. Every 10 iterations, the RNN parameters are updated with a learning rate of $10^{-7}$, while freezing the remainder of the network parameters. As before, the label estimation and training steps are repeated until convergence.%\textcolor{red}{Since the network is learned from scratch and we use a batch size of 4 unlike \cite{DBLP:conf/iccv/0001JRVSDHT15}, we need a much higher learning rate for the network than they used. However, we observed that the RNN parameters are more sensitive than weights of the CNN. Therefore, we opted for using two optimizing stages with different learning rates. The idea behind optimizing the network and RNN in an alternating fashion is giving network time to adapt to the current state of the RNN module.} 

\subsection{Quantifying segmentation uncertainty}\label{sec:uncertainty}

In order to prevent segmentation errors from early recursions from propagating we investigate the following strategy to reset labels predicted with insufficient certainty after each E step. We add dropout with probability 0.5 to the 5 innermost blocks of our U-Net architecture during training. In order to estimate the new optimal labeling $\mathbf{z^*}$ we perform 50 forward passes with dropout similar to \cite{DBLP:journals/corr/KendallBC15}. Rather than a single output this yields a distribution of logits and softmax outputs for each pixel and label. We then compare the logits distributions of the label with the highest and second highest softmax mean for each pixel using a Welch's t-test. If the logits come from a distribution with the same mean with $p\geq0.05$ we conclude that the label was not predicted with sufficient certainty and reset its labeling to ``uncertain''. Thus, in the subsequent M-step the network will be free to predict any label in that location. Otherwise, we set the pixel to the label with the highest probability. %We found that mostly pixels at border regions get reset, which is desirable. 

\section{Experiments and Results}
\label{sec:experiments}

We trained and evaluated the methods on two publicly available datasets: the ACDC cardiac segmentation challenge data \cite{bernard2018deep} for which the Myocardium (Myo), the left and right ventricles (LV and RV) have been annotated, and the NCI-ISBI 2013 prostate segmentation challenge data \cite{bloch2015nci} for which reference annotations of the central gland (CG) and the peripheral zone (PZ) were available. For the cardiac data we split the data into 160 training volumes and 40 validation volumes, and evaluated the algorithms on 100 images using the challenge server. For the prostate data we split 29 available training volumes into 12 training, 7 validation and 10 testing volumes. Training was performed on 2D slices. 

We used $\tau=0.99$ for the cardiac and $\tau=0.90$ for the prostate experiments. For the separate CRF we used $w_1=5, w_2=10, \sigma_\alpha=2, \sigma_\beta=0.1, \sigma_\gamma=5$ for the cardiac experiments and $w_1=6, w_2=10, \sigma_\alpha=3, \sigma_\beta=0.01, \sigma_\gamma=2, \tau=0.9$ for the prostate, and for the CRF-RNN we used $\sigma_\alpha=160$, for the cardiac data, $\sigma_\alpha=250$ for the prostate, and $\sigma_\beta=3, \sigma_\gamma=10$ for both datasets. 

In the following experiments, the simple recursive training strategy which does not make use of pairwise terms in Eq. \ref{eq:CRF}, nor uncertainty estimation, is called \emph{base}. We evaluated the performance with and without the components discussed above. Additionally, we also investigated the same segmentation architecture on the fully labeled data to obtain an upper bound on the performance, and a version of \emph{base} in which we did not perform any recursions, but used the network parameters learned directly on the seed regions $\hat{\mathbf{z}}$.  %\footnote{The differences in the scale of $\sigma_\alpha$ are due to different implementations.}

%\textcolor{blue}{Surprisingly, Method 1 yielded slightly worse results when applying the CRF for the final prediction. We thus report the results without CRF for the final prediction. We believe, this might be due to the hyperparameters being sub-optimally determined, and will investigate this further in future work.}

\begin{figure}[t]\label{fig:qual}
\centering
\includegraphics[width=0.92\textwidth]{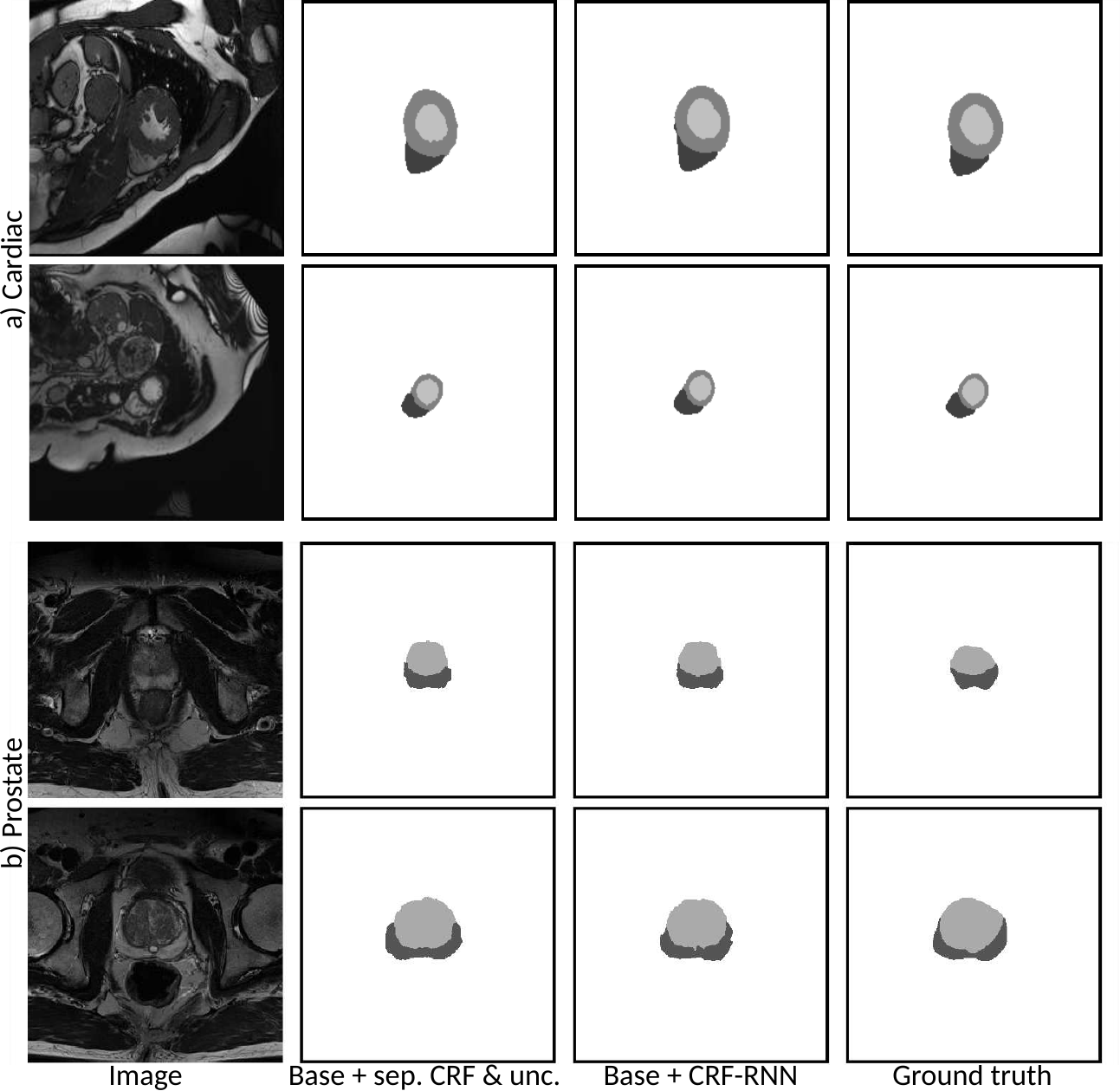}
\caption{Randomly sampled example segmentations for the two best performing training strategies for the a) cardiac and b) prostate data.}
\end{figure}

The Dice scores with respect to the reference annotations for all the examined methods and structures are shown in Table \ref{tab:MethodComparison}. Note that ACDC challenge server did not allow for higher precision Dice reporting in the post-challenge phase. Example segmentations for the two best performing methods are shown in Fig. \ref{fig:qual} for the cardiac and prostate data, respectively. 

% \begin{table*}[th!]
% \caption{Dice scores on Cardiac and Prostate datasets.}
% \label{tab:MethodComparison}
% \centering
% \begin{tabular}{l | cccc | ccc}
% &\multicolumn{4}{|c|}{\textbf{Cardiac dataset}}&\multicolumn{3}{|c}{\textbf{Prostate dataset}} \\
%  & LV  & RV  & Myo & Avg & PZ & CG & Avg \\
% \hline
% Fully sup.                   & 0.935 & 0.905 & 0.895 & 0.912 & 0.746 & 0.889 & 0.818  \\
% \hline 
% Method 1 (sep. CRF)          & 0.910& \textbf{0.890} & 0.840 & \textbf{0.880} & \textbf{0.722} & 0.839 & 0.780\\
% Method 2 (CRF-RNN)           & \textbf{0.915} &0.885 & 0.840 & \textbf{0.880} & 0.698 & \textbf{0.863} & \textbf{0.781}\\
% \hline 
% Method 1 (no CRF)            & 0.910 & \textbf{0.890} & 0.840 & \textbf{0.880} & 0.720 & 0.837 & 0.778\\
% Method 1 (no unc.)           & 0.890 & 0.880 & 0.840 & 0.870 & 0.698 & 0.837 & 0.767 \\
% Method 1 (no CRF \& no unc.) & 0.905 & 0.880 & 0.835 & 0.873 & 0.670 & 0.829 & 0.750 \\
% \hline
% \end{tabular}
% \end{table*}

\begin{table*}[th!]
\caption{Dice scores on Cardiac and Prostate datasets.}
\label{tab:MethodComparison}
\centering
\begin{tabular}{l | ccc |c | cc| c}
&\multicolumn{4}{|c|}{\textbf{Cardiac dataset}}&\multicolumn{3}{|c}{\textbf{Prostate dataset}} \\
 & LV  & RV  & Myo & Avg & PZ & CG & Avg \\
\hline
Base (no recursion)     & 0.895 & 0.875  & 0.825 & 0.865 & 0.631 & 0.827  & 0.729 \\
Base                    & 0.905 & 0.880  & 0.835 & 0.873 & 0.670 & 0.829  & 0.750 \\
Base + separate CRF         & 0.890 & 0.880  & \textbf{0.840} & 0.870 & 0.698 & 0.837  & 0.767 \\
Base + CRF-RNN          & \textbf{0.915} & 0.885 & \textbf{0.840} & \textbf{0.880} & 0.698 & \textbf{0.863} & \textbf{0.781} \\
Base + uncertainty      & 0.910 & \textbf{0.890} & \textbf{0.840} & \textbf{0.880} & 0.720 & 0.837 & 0.778 \\
Base + sep. CRF \& unc. & 0.910 & \textbf{0.890} & \textbf{0.840} & \textbf{0.880} & \textbf{0.722} & 0.839 & 0.780\\
Base + CRF-RNN \& unc. & \textbf{0.915} & 0.885 & \textbf{0.840} & \textbf{0.880} & 0.710 & 0.834 & 0.772 \\
\hline
Fully supervised        & 0.935 & 0.905 & 0.895 & 0.912 & 0.746 & 0.889 & 0.818  \\
\end{tabular}
\end{table*}

We observe that a) the recursive training regime led to substantial improvements over non-recursive training, b) the dropout based uncertainty was responsible for the largest improvements, c) additional CRF led to further, albeit smaller improvements, d) using CRF-RNN without uncertainty led to similar results as the separate CRF with uncertainty, e) applying dropout uncertainty in conjunction with the CRF-RNN did not lead to additional improvements and performed slightly worse on the prostate. We believe this is due to the CRF-RNN module leading to unusual logit distributions at its input. On average, the training frameworks with 1) CRF-RNN, and with 2) separate CRF and uncertainty performed the best and similar to each other. Future work on integrating uncertainty with the CRF-RNN may lead to further improvements. 

Most importantly, the results show that our proposed training strategy allows to learn a pixel-level segmentation network using scribble supervision alone with a remarkably small degradation compared to the fully supervised upper bound. For instance, the performance of the CRF-RNN method is only 4.5\% worse on the prostate, and 2.9\% worse on the cardiac data compared to fully supervised training. These results are also confirmed by the qualitative analysis. We believe this is likely an acceptable error margin for certain quantification studies where precise border delineation is of secondary importance such as automatic estimation cardiac ejection fractions \cite{bernard2018deep}. %We observed that in this case the logits before the CRF-RNN were activated by more than just the desired structures, appearing to produce candidate regions for the CRF.

% \begin{figure}[t]\label{fig:qual_cardiac}
% \centering{}
% \includegraphics[width=1\textwidth]{images/heart_results.eps}
% \caption{Example cardiac segmentations for Method 1 and Method 2.}
% \end{figure}

% \begin{figure}[t]\label{fig:qual_prostate}
% \centering
% \includegraphics[width=1\textwidth]{images/pro_results.eps}
% \caption{Example prostate segmentations for Method 1 and Method 2.}
% \end{figure}

\section{Conclusion}

In this paper, we investigated training strategies to train a fully automatic segmentation network with scribble supervision alone. We demonstrated the feasibility of the techniques on two publicly available medical image datasets and showed that only a remarkably small performance degradation is incurred with respect to fully supervised upper bound networks. %, but requiring only a fraction of the annotation time. 

\subsubsection*{Acknowledgements}
This work was partially supported by the Swiss Data Science Center. One of the Titan X Pascal used for this research was donated by the NVIDIA Corporation.

\bibliographystyle{splncs04}
\bibliography{references}

\end{document}